\documentclass{article}
\usepackage{spconf,amsmath,graphicx}
\usepackage{soul}
\usepackage{subfigure}
\usepackage{bm}
\usepackage{amsmath}
\usepackage{amssymb}
\usepackage{tablefootnote}
\usepackage{multirow}
\usepackage{booktabs}
\usepackage{arydshln}
\usepackage{color}
\usepackage{tikz}
\usepackage{CJKutf8}
\usepackage{hyperref}
\usepackage{pifont}
\usepackage{marvosym}
\usepackage{fontawesome}

\title{CPPF: A contextual and post-processing-free model for automatic speech recognition}
\name{Lei Zhang$^{1~*}$\thanks{$^*$Work was done during the internship at Meituan.}, Zhengkun Tian$^{2}$, Xiang Chen$^2$, Jiaming Sun$^2$, Hongyu Xiang$^2$, Ke Ding$^2$, Guanglu Wan$^2$}
\address{$^1$School of Computer Science and Technology, Soochow University, China\\ $^2$Meituan}
\begin{document}
\ninept
\maketitle
\begin{CJK}{UTF8}{gkai}
    
\begin{abstract}
ASR systems have become increasingly widespread in recent years. However, their textual outputs often require post-processing tasks before they can be practically utilized. To address this issue, we draw inspiration from the multifaceted capabilities of LLMs and Whisper, and focus on integrating multiple ASR text processing tasks related to speech recognition into the ASR model. This integration not only shortens the multi-stage pipeline, but also prevents the propagation of cascading errors, resulting in direct generation of post-processed text. In this study, we focus on ASR-related processing tasks, including Contextual ASR and multiple ASR post processing tasks. To achieve this objective, we introduce the CPPF model, which offers a versatile and highly effective alternative to ASR processing. CPPF seamlessly integrates these tasks without any significant loss in recognition performance.
\end{abstract}

\vspace{-1em}

\section{Introduction}
\label{sec:introduction}
\begin{figure*}[t!]
\centering
\includegraphics[width=0.94\textwidth]{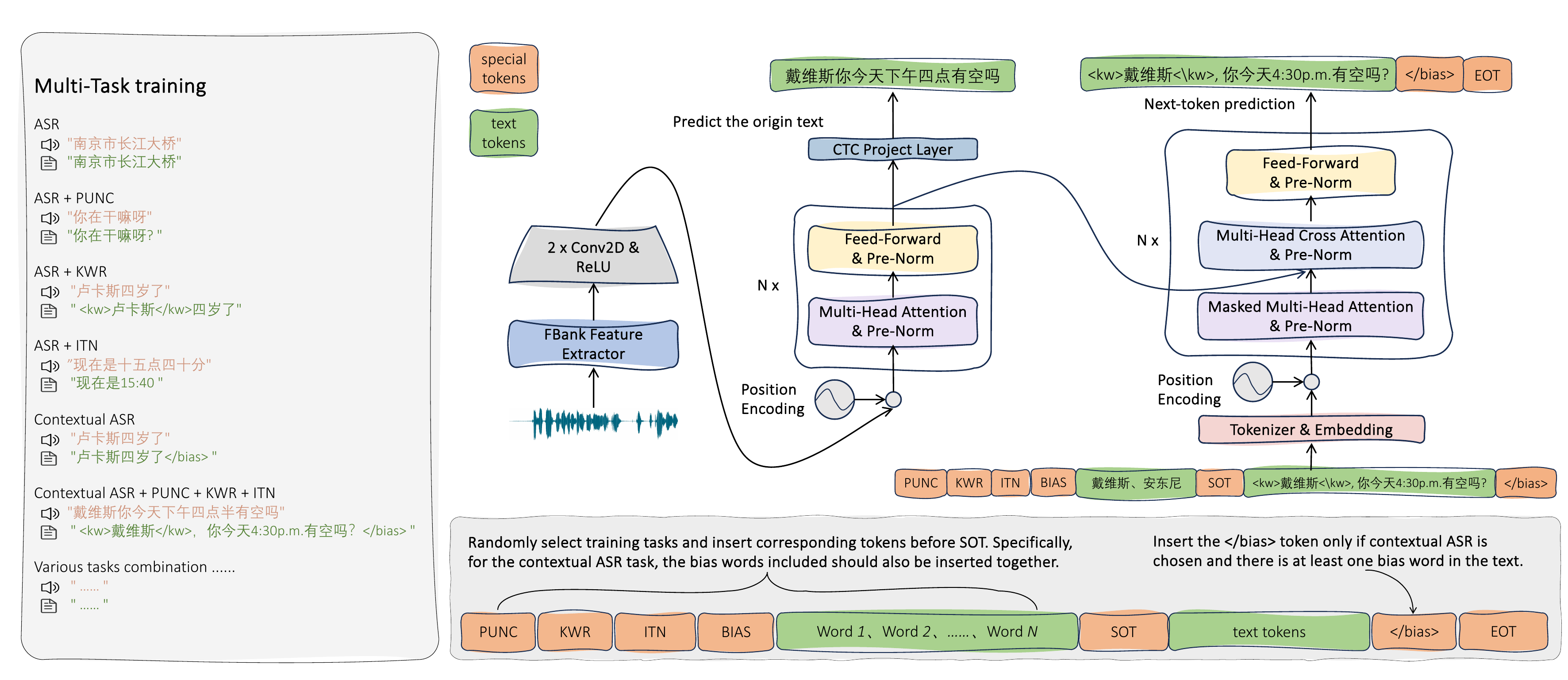}
\caption{CPPF overview. 
The Transformer model, equipped with this CTC module, has undergone training across a spectrum of ASR-related tasks, including ASR, contextual ASR, punctuation restoration, key word recognition, and inverse text normalization. 
These tasks are jointly represented as a sequence of tokens for prediction by the decoder, shorten the pipeline of multiple stages in the traditional ASR post-processing. The multitask training format incorporates special tokens, which act as task specifiers, as elaborated in Section \ref{sec:multitask_format}.}
\label{img:model}
\end{figure*}

With the rapid development of Large Language Models (LLMs) \cite{ouyang2022training,chowdhery2022palm,touvron2023llama,OpenAI2023GPT4TR}, such as ChatGPT, we are witnessing the emergence of a technological revolution. These models possess a broad spectrum of capabilities for text-based operations, including tasks such as text generation and editing, all guided by a diverse array of instructive texts. As a result, they have demonstrated their versatility across various task domains. Automatic Speech Recognition (ASR) system yields outputs that lack punctuation and are often subject to normalization, leading to suboptimal readability \cite{pratap2020massively,babu2021xls}. Common post-processing tasks such as Punctuation Restoration (PUNC) and Inverse Text Normalization (ITN) are frequently employed to enhance the comprehensibility of these outputs. Furthermore, there are instances where further natural language understanding tasks are required to be performed on these output, leveraging the transcribed text to extract valuable information, such as Named Entity Recognition (NER), word segmentation, and information extraction.

Notably, the widely adopted attention-based encoder-decoder architecture in ASR models can be viewed as an autoregressive language model. Inspired by the remarkable multifunctionality of LLMs, there is a growing interest in expanding the language comprehension capabilities of ASR models. To achieve this, we propose integrating text processing tasks regarding speech recognition into the ASR model, consolidating them into the decoder module. With knowledge of multiple task types, the decoder can directly decode and generate the final output. This approach not only shortens the entire ASR pipeline, but also mitigates cascading errors that can be introduced by multi-stage pipeline processing.

In the field of speech recognition, several endeavours have emerged to integrate multiple tasks into a single model, including notable efforts such as Whisper \cite{radford2023robust}, WavPrompt \cite{gao2022wavprompt}, SpeechPrompt \cite{chang2022speechprompt}, and SpeechPrompt v2 \cite{chang2023speechprompt}. Whisper focuses on various tasks initiated directly from speech data, employing a standard encoder-decoder Transformer architecture for training across a diverse set of tasks, including multilingual speech recognition, speech translation, spoken language identification, and voice activity detection. WavPrompt, SpeechPrompt, and SpeechPrompt v2 focus on speech classification and detection tasks. WavPrompt utilizes pre-trained wav2vec 2.0 \cite{baevski2020wav2vec} and GPT-2 \cite{radford2019language} models, directly concatenating speech encoder and text embedding information. It generates responses based on the input text prompts. In contrast, SpeechPrompt and SpeechPrompt v2 utilize the HuBERT model and prompt tuning with a prompt module on various speech classification and detection tasks. However, it is essential to note that these models only focus on recognition, translation, classification, and detection tasks originating directly from speech data, without giving attention to a range of text-related tasks associated with speech processing.

Furthermore, Ghannay et al. (2018) \cite{ghannay2018end}, Yadav et al. (2020) \cite{yadav20b_interspeech}, and Chen et al. (2022) \cite{chen2022aishell} propose an approach to directly obtain Named Entity (NE) information from speech while performing speech recognition tasks. This approach eliminates the need for a two-step process where speech is first transcribed and then text-based NER is applied. Instead, they integrate special tags into the transcription text to denote NE, facilitating the direct extraction of NE information from speech using ASR model.
Similarly, Kim et al. (2023) \cite{kim23_interspeech} have proposed a method that directly produces text with punctuation during recognition.
However, it is worth noting that they focus exclusively on a single specific task, and their methods are engineered in such a way that each model is limited to performing only that particular function. When these models are trained with the objective sequence tailored for this specific task, they are constrained to predict target sequences corresponding to this specific task, without the flexibility to decide whether or not to perform this task.

In this paper, our focus is on text processing tasks that are intimately linked to speech recognition, and we consolidate these responsibilities within the decoder module. This strategic approach empowers the decoder to gather knowledge from various tasks, facilitating the generation of task-specific results while simultaneously processing speech information. We introduce \textbf{CPPF}, an innovative ASR model that stands for \textbf{C}ontextual and \textbf{P}ost-\textbf{P}rocessing-\textbf{F}ree model. CPPF marks a paradigm shift in the ASR landscape as it integrates knowledge from a variety of tasks directly into the decoder. This expansion not only enhances the transcription performance of the ASR model, but also equips it with capabilities across a spectrum of text processing tasks associated with speech recognition. In this paper, our focal point covers four key tasks: Contextual ASR, PUNC, Key Word Recognition (KWR), and ITN. These tasks are fundamental and prevalent in the field of text processing related to speech recognition. CPPF seamlessly integrates these tasks, allowing it to produce text results that go through these task-specific processes while simultaneously recognizing speech. Significantly, CPPF shortens the ASR processing pipeline, eliminating the cascading errors often associated with multi-stage pipeline processing. This innovative model offers a more efficient and direct solution for the integration of ASR and text processing.
\vspace{-1em}
\section{Method}
\label{sec:method}
\vspace{-1em}
\subsection{Model}
The architectural structure of the CPPF is presented in Figure \ref{img:model}. We employ the well-established Transformer architecture proposed by Vaswani et al. (2017) \cite{vaswani2017attention} for our research. As for the encoder, it can also be replaced with the Conformer \cite{gulati2020conformer}, which has been widely demonstrated to offer superior performance. Distinguishing itself from the conventional Transformer, we introduce the CTC \cite{graves2006connectionist} module to enhance the encoder's understanding of speech. Notably, irrespective of the specific combination of various tasks the model performs, the CTC's target sequence consistently aligns with the original ASR text. For model training, we employ a mask matrix to mask the decoder output sequence positions before \texttt{<|SOT|>}, focusing solely on the decoder's output between \texttt{<|SOT|>} and \texttt{<|EOT|>}.

For speech input, we employ 80-dimensional Fbank features with a frame shift of 10ms and a window length of 25ms. 
To normalize these features, we globally scale the input values to lie within the range of -1 to 1 while maintaining an approximately zero mean across the training dataset. 
Subsampling is implemented through two Conv2D layers with filter widths of 3 and a stride of 2, accompanied by a ReLU activation function.

In text processing, we employ a character-level tokenizer with a vocabulary size of 5,390.
\vspace{-1em}
\subsection{Contextual ASR and ASR Post-Processing Tasks}
\vspace{-0.5em}
\label{sec:tasks}
In this work, we focus on four pivotal text-processing tasks pertaining to ASR: Contextual ASR, PUNC, KWR, and ITN.
\vspace{-1em}
\subsubsection{Contextual ASR}
\label{sec:contextual_asr}
Contextual ASR is a widely used text processing technique in ASR, primarily employed to enhance the recognition of bias words. In this work, contextual ASR takes a significant departure from traditional paradigms. Instead of calculating bias at each prediction step like CLAS \cite{pundak2018deep}, we directly insert the bias word list, joined by a special token \texttt{<|separator|>}, into the decoder's initial tokens. 
Regarding the target sequence, we also employ a different approach from CLAS. Instead of adding a special token \texttt{</bias>} after each bias word, we only append at the end of sentences containing bias words to minimize the impact of \texttt{</bias>} tokens on the output sequence.
This approach eliminates the necessity for an additional bias encoder module, while capitalizing on the inherent language modeling capabilities of the decoder. As a result, this innovation greatly improves the speed of contextual ASR, although there are limitations on the number of bias words that can be accommodated.

For the bias word list, we construct a named entity set from the training set. For each sample, we randomly select a subset of words from current sentence words and the named entity set to create the current contextual bias word list. For the test set, we establish a fixed test set, where the contextual bias word lists are randomly selected from words in the current sentence and named entities of the test set. In this work, we employ the Hanlp \cite{he-choi-2021-stem} toolkit utilizing the ner\_bert\_base\_msra\footnote{file.hankcs.com/hanlp/ner/ner\_bert\_base\_msra\_20211227\_114712.zip} model for named entities extraction and the pkuseg\footnote{https://github.com/lancopku/pkuseg-python} \cite{pkuseg} toolkit for Chinese word segmentation extraction.
\vspace{-1em}
\subsubsection{Key Word Recognition}
For ASR output, key information extraction and highlighting are often required, and NER is a common approach to this purpose. In our work, we extend NER to better align with practical ASR application scenarios. Instead of focusing exclusively on traditional named entities, we identify and emphasize all the key words relevant to real-world applications. In this work, we employ all named entities with their categories removed as key words. In practical applications, we retain the flexibility to extend the set of key words beyond named entities. This extension expands the applicability of our approach, making it suitable for a wider range of application scenarios.

In this study, to highlight key words, we employ a pair of special tokens, \texttt{<kw>} and \texttt{</kw>}, to directly mark keywords within the text. Consistent with the contextual ASR task, using the Hanlp toolkit to extract named entities.
\vspace{-1em}
\subsubsection{Punctuation Restoration}
PUNC task is one of the most commonly used post-processing tasks in ASR. ASR outputs typically consist of normalized text without punctuation, and for improved readability, the PUNC task is necessary to add punctuation. Similar to widely-used PUNC tasks in Chinese, we focus on three primary punctuation marks: ``\texttt{，}'', ``\texttt{。}'', and ``\texttt{？}''. For the PUNC data obtained, we utilize an in-house model. This model was trained using the 256-dimensional, 8-layer RoBERTa\footnote{https://huggingface.co/uer/chinese\_roberta\_L-8\_H-128} \cite{liu2019roberta} architecture provided by the UER-py \cite{zhao2019uer} release.
\vspace{-1em}
\subsubsection{Inverse Text Normalization}
ITN is another commonly used post-processing task in ASR. ITN can significantly enhance the readability of ASR output, for instance, transforming ``one hundred fourteen thousand five'' into ``114,005'' thereby greatly improving the text's readability. For ITN transformation, we utilize the WeTextProcessing\footnote{https://github.com/wenet-e2e/WeTextProcessing} toolkit, which is an open-source and user-friendly TN/ITN tool specifically designed for the Chinese language. This toolbox employs a WFST-based \cite{ebden2015kestrel} approach built on grammar rules, making it a reliable choice for ITN. 
\vspace{-1em}
\subsection{Multitask Format}
\label{sec:multitask_format}
We enable the flexible integration of $0$ to $N$ text-processing tasks related to ASR, with $N$ representing the allowable number of post-processing tasks. In this work, we configure $N$ to be $4$, encompassing Contextual ASR, PUNC, KWR, and ITN tasks. During the training phase, we allocate probabilities for the selection of each task. In this work, Contextual ASR has a 50\% chance of being selected, while PUNC, KWR, and ITN each have a 30\% chance. When a task is selected, its corresponding special label is added to the initial token list. Especially, when Contextual ASR is selected, we need to insert the bias word list, joined by a special token \texttt{<|separator|>}.

Upon the completion of adding all special labels for tasks, we introduce a \texttt{<|SOT|>} label, followed immediately by the insertion of the output corresponding to the current post-processing task combination. 
Especially, if the contextual ASR task is selected and words from the bias word list are present in the target text, we append a \texttt{</bias>} label at the end of the output text.

\section{Experiments}
\label{sec:experiments}
\begin{table*}
    \begin{small}
    \centering
    \begin{tabular}{lccccccccccc}
    \toprule
    \multirow{2}*{Model} & ASR & \multicolumn{3}{c}{PUNC} & \multicolumn{3}{c}{KWR} & ITN & \multicolumn{2}{c}{Contextual ASR} \\
    \cmidrule(lr){2-2}
    \cmidrule(lr){3-5}
    \cmidrule(lr){6-8}
    \cmidrule(lr){9-9}
    \cmidrule(lr){10-11}
    ~  & $\downarrow$CER  & $\uparrow$P & $\uparrow$R & $\uparrow$F1 &  $\uparrow$P    & $\uparrow$R & $\uparrow$F1 & $\uparrow$SA & $\downarrow$CER  & $\uparrow$R \\
    Baseline & 7.08 & - & - & - & - & - &  -    & - & -    & - \\
    ~~ w/ pipeline post-processing  & -    & \textbf{91.50} & \textbf{91.87} & \textbf{91.68} & \textbf{73.54} & \textbf{76.49} & \textbf{74.99} & 61.46 & - & - \\
    E2E speech NER & - & - & - & - & 66.29 & 67.63 & 66.95 & - & - & - \\
    CPPF & \textbf{6.92} & 84.05 & 82.72 & 83.38 & 67.15 & 69.86 & 68.48 & \textbf{61.92} & \textbf{4.71} & \textbf{96.57} \\
    \midrule
    Conformer Encoder \\ 
    Baseline & 6.40  & - & - & - & - & - &  -    & - & -    & -  \\
    ~~ w/ pipeline post-processing  & -    & \textbf{91.97} & \textbf{92.53} & \textbf{92.25} & \textbf{75.84} & \textbf{77.18} & \textbf{76.50} & 64.00 & - & - \\
    CPPF & \textbf{6.33} & 85.09 & 84.12 & 84.60 & 70.77 & 71.58 & 71.17 & \textbf{64.12} & \textbf{4.56} & \textbf{96.49} \\

    \midrule
    CPPF & 6.92    & 84.05    & 82.72    & 83.38    & 67.15    & 69.86    & 68.48 & \textbf{61.92} & \textbf{4.71} & \textbf{96.57} \\
    ~~ w/o PUNC & 7.19    & - & - & - & 66.44    & \textbf{70.22} & 68.28 & 60.82 & 4.77 & 96.42 \\
    ~~ w/o KWR  & 6.98    & \textbf{84.42} & \textbf{82.87} & \textbf{83.64} & - & - & - & 61.52 & 4.86 & 96.28 \\
    ~~ w/o ITN  & \textbf{6.88} & 84.00    & 82.50    & 83.24    & \textbf{67.38} & 69.86    & \textbf{68.60} & - & 4.73 & 96.40 \\
    ~~ w/o Contextual ASR & 6.89    & 84.26    & \textbf{82.87} & 83.56    & 65.83    & 69.86    & 67.78 & 61.90 & - & - \\

    \bottomrule 
    \end{tabular}
    \caption{Results of experiments. Experiment results are categorized using horizontal dividers. 
    }
    \label{table:experiments}
    \end{small}
\end{table*}

\subsection{Implementation Details}
\label{sec:imple_details}
Experiments are conducted on Aishell-2 \cite{du2018aishell}. Aishell-2 comprises 1,000 hours of Chinese spoken data, all sampled at a 16 kHz rate.

We exclusively utilize a compact model with the following parameter configuration: a hidden layer dimension of 256, an Encoder comprising 4 attention heads and 12 layers, and a Decoder consisting of 4 attention heads and 6 layers. 

For training process, we adopted a learning rate of 2e-3, integrated a warm-up scheduler comprising 16,000 steps, and employed the Adam optimizer. Our training continued until we observed no significant improvement in the model's performance on the validation dataset for 30 consecutive epochs, which we considered an indicator of robust convergence on our training data. Additionally, we set the batch size to 64 in 8 GPU parallel training. The implementation of our model is based on the Wenet \cite{yao2021wenet} framework.

For evaluation process, we utilized the average model weights from last 30 epochs. We only employed the results from decoder.
\vspace{-1em}
\subsection{Metric}

Regarding evaluation metrics, we align with the mainstream standards for each task. We employ the Character Error Rate (CER) for the ASR task for evaluation. For both PUNC and KWR tasks, we utilize Precision (P), Recall (R), and F1-score (F1) as our evaluation metrics. 
For the ITN task, we evaluate using Sentence Accuracy (SA). 
We combine CER with Bias Word Recall (R) for the Contextual ASR task for performance evaluation. 
\vspace{-1em}

\subsection{Experiments Settings}
\label{sec:experiments_settings}
\noindent\textbf{Baseline.} 
We employed the fundamental encoder-decoder Transformer with CTC module based on the Wenet, keeping the model parameters and training strategy consistent with the CPPF model.

\noindent\textbf{Baseline with pipeline post-processing.} 
We utilized the ASR transcripts obtained from the baseline setup, followed the processing methodology outlined in Section \ref{sec:tasks} to obtain the corresponding results, which were subsequently subjected to evaluation.

\noindent\textbf{E2E speech NER.} We implemented the methods introduced by Chen et al. (2022) \cite{chen2022aishell} to allow a comparative analysis of previous E2E approaches. The experimental setup, including parameters, dataset, and evaluation metrics, is consistent with KWR.

\noindent\textbf{CPPF.} 
We trained the CPPF model by utilizing various combinations of Contextual ASR, PUNC, KWR, and ITN tasks, as outlined in detail in Section \ref{sec:tasks}. The implementation details of CPPF are described in Section \ref{sec:imple_details}.
\vspace{-1em}
\subsection{Experiments Results}
\begin{table}
    \begin{small}
    \centering
    \setlength{\tabcolsep}{4pt}
    \begin{tabular}{lcccccccc}
    \toprule
    \multirow{2}*{Model} & \multicolumn{3}{c}{PUNC} & \multicolumn{3}{c}{KWR} & ITN  \\
    \cmidrule(lr){2-4}
    \cmidrule(lr){5-7}
    \cmidrule(lr){8-8}
    ~    & $\uparrow$P     & $\uparrow$R     & $\uparrow$F1 &  $\uparrow$P    & $\uparrow$R     & $\uparrow$F1 & $\uparrow$SA \\
    Pipeline  & \textbf{91.86} & \textbf{92.05} & \textbf{91.95} & \textbf{75.50} & \textbf{77.13} & \textbf{76.31} & \textbf{62.19}   \\
    CPPF  & 83.78 & 82.30 & 83.03 & 67.99 & 70.39 & 69.17  & 61.68  \\

    \bottomrule 
    \end{tabular}
    \caption{Results of multiple task. }
    \label{table:multiple_task}
    \end{small}
\end{table}

The experimental results presented in Table \ref{table:experiments} unequivocally demonstrate CPPF's proficiency across various post-processing tasks. Notably, CPPF even shows a slight improvement in ASR task performance. In the KWR task, CPPF outperforms the E2E Speech NER model, achieving a higher F1 score of 1.53. However, when compared to pipeline methods, CPPF falls behind by 6.51 points in terms of F1 score. This discrepancy can be attributed to pipeline post-processing methods that benefit from pre-trained language models, which CPPF does not employ. These findings are consistent with those reported by Chen et al. (2022) \cite{chen2022aishell}. The PUNC task, similar to the KWR task, relies on pre-trained language models in pipeline processing, resulting in superior performance compared to CPPF. Nevertheless, CPPF still delivers commendable performance. In the ITN task, where pipeline post-processing methods do not use pre-trained language models, CPPF exhibits slightly higher performance. Additionally, CPPF exhibits contextual ASR capabilities, achieving a CER of 4.71 and a recall rate of 96.57\% for bias words when contextual ASR is utilized.

Furthermore, CPPF has the capability to perform combinations of multiple tasks. For instance, it can directly predict the final results of task combinations such as PUNC+KWR+ITN. Experimental results are presented in Table \ref{table:multiple_task}, where Pipeline represents the output obtained by sequentially applying tasks to the baseline. CPPF's unique ability to directly derive the final results of multiple task combinations without the need for sequential execution sets it apart from traditional pipeline methods. Despite pipeline methods using pre-trained language models, which CPPF does not employ, CPPF consistently delivers commendable performance across various tasks. In addition, it streamlines the processing pipeline by directly predicting the results of multiple task combinations, highlighting its efficiency and effectiveness.
\vspace{-1em}

\subsection{Impact with Encoder Module}
To investigate the impact of different encoder architectures on CPPF, we replaced the encoder component with a Conformer \cite{gulati2020conformer} architecture for experimentation while keeping all other parameters constant. The experimental results, shown in the second main row of Table \ref{table:experiments}, illustrate a significant improvement in both baseline, baseline with pipeline post-processing and CPPF across all tasks. It is noteworthy that after the improvement of encoder performance, the performance gap between CPPF and the pipeline approach in text processing tasks has been significantly reduced. 
This trend indicates that in the future, if we continue to improve the performance of the encoder or integrate pre-trained models into CPPF, we have the potential to outperform the pipeline approach in various aspects, including speed and performance. 
\vspace{-1em}

\subsection{Impact with Different Task}
To evaluate the impact of varying the number of text processing tasks on CPPF, we conducted experiments involving the selective removal of individual tasks. Experimental results are presented in the third main row of Table \ref{table:experiments}. Experimental results show that when one task is removed, the performance of the remaining tasks experiences only minor fluctuations within a narrow range. This observation highlights the adaptability of CPPF, as it can seamlessly accommodate adding or removing various tasks.

In future work, we aim to expand the number of tasks to enrich the model's knowledge across various tasks. In addition, we will delve deeper into understanding the effects of introducing or removing tasks on existing tasks. For instance, the experiment where we removed the PUNC task revealed significant fluctuations in ASR performance. This phenomenon motivates us to explore the deep relationships between various tasks in future research.
\vspace{-1em}
\section{Limitations}
\label{sec:Limitations}
Our presented work may see further improvements in addressing two potential limitations.

Firstly, in contrast to pipeline methods, we have not integrated pre-trained language models into CPPF, which results in a disadvantage in text processing performance. In our future work, we intend to explore integrating pre-trained language models into CPPF to achieve consistent or even enhanced text processing performance.

Secondly, for specific tasks requiring additional input, such as Contextual ASR, CPPF's approach of inserting them directly into the decoder's initial tokens limits the length of the additional input. However, this approach also offers advantages, such as avoiding the need to compute the bias encoder at each step, as required in CLAS. In the future, we need to further enhance our approach to accommodate longer inputs for tasks such as Contextual ASR.

Despite these limitations, CPPF continues to deliver commendable performance and significantly shortens the ASR processing pipeline. It can directly output the final results of multiple task processes without needing sequential processing.
\vspace{-1em}
\section{Conclusion}
\label{sec:conclusion}

This paper presents the CPPF model, which offers a simple yet highly effective approach, representing an important milestone in the integration of ASR and multiple text-processing tasks within a unified framework. CPPF can directly generate results that have undergone multiple post-processing tasks, simplifying the ASR pipeline and reducing the propagation of consequential errors that often occur in the cascading stages of the pipeline process. Furthermore, CPPF produces text-processed textual outputs. Extensive experiments on the Aishell-2 dataset validate the effectiveness of our proposed solution. In summary, CPPF provides a comprehensive and promising solution for seamlessly integrating ASR and multiple text-processing tasks within a unified model.
\newpage
\bibliographystyle{IEEEbib}
\bibliography{strings,refs}

\end{CJK}

\end{document}